\documentclass{article}
\usepackage{spconf,amsmath,graphicx, epstopdf}
\usepackage{caption}
\usepackage{subcaption}
\usepackage{multirow}
\usepackage{tablefootnote}
\usepackage{cite}
\usepackage{tabularx}

\title{CHARACTERIZATION OF DIM LIGHT RESPONSE IN DVS PIXEL: DISCONTINUITY OF EVENT TRIGGERING TIME}

\name{Xiao Jiang$^1$, Fei Zhou$^{1,2,3,4,*}$\thanks{* Corresponding author.}}
\address{1. College of Electronic and Information Engineering, Shenzhen University, Shenzhen, China\\ 2. Guangdong Key Laboratory of Intelligent Information Processing, Shenzhen, China\\ 3.  Guangdong-Hong Kong Joint Laboratory for Big Data Imaging and Communication, China\\ 4. Peng Cheng Laboratory, Shenzhen, China}
\begin{document}
%
\maketitle
\begin{abstract}
Dynamic Vision Sensors (DVS) have recently generated great interest because of the advantages of wide dynamic range and low latency compared with conventional frame-based cameras. However, the complicated behaviors in dim light conditions are still not clear, restricting the applications of DVS. In this paper, we analyze the typical DVS circuit, and find that there exists discontinuity of event triggering time. In dim light conditions, the discontinuity becomes prominent. 
We point out that the discontinuity depends exclusively on the changing speed of light intensity.  Experimental results on real event data validate the analysis and the existence of discontinuity that reveals the non-first-order behaviors of DVS in dim light conditions. 
\end{abstract}
\section{Introduction}
\label{sec:intro}

Dynamic vision sensors (DVS) detect temporal contrast in light intensity, different from conventional cameras sensing light intensity directly \cite{delbruck1994adaptive, delbruck2012icip, delbruck2022icip, delbruck2008IEEEJSSC, delbruck2022IEEETPAMI}. It imitates human retina, a logarithmic and first-order system sensitive to a fixed light contrast but invariant to absolute light intensity \cite{mahowald1991silicon}. The humanoid behaviors facilitate the high dynamic range (more than 6 decades), low latency, and energy saving \cite{indiveri2011neuromorphic}. 

However, the first-order response does not suit for the DVS in dim light conditions, due to the properties of DVS circuit\cite{delbruck2008IEEEJSSC}. The imperfect behaviors in dim light conditions have been studied by researchers. To increase the dynamic range, Delbruck and Mead\cite{delbruck1994adaptive} suppress the non-first-order response in the conventional logarithmic light receptor by an adaptive element and cascode feedback loop. This adaptive photoreceptor invented by Delbruck and Mead is the prototype of recent DVS \cite{muglikar2021calibrate, brewer2021comparative, mueggler2017event, gehrig2020video}. Lichtsteiner, Posch, and Delbruck \cite{delbruck2008IEEEJSSC} point out that the first-order response is contaminated by the photodiode parasitic capacitor and resistance of the feedback metal-oxide-semiconductor field-effect transistor (MOSFET) without further analysis. Hu, Liu and Delbruck \cite{hu2021v2e} design a widely-used DVS simulator, which forms the operation of DVS in dim light conditions as an infinite impulse response (IIR) low pass filter. Graca and Delbruck \cite{graca2021unraveling} research on the shot noise of DVS in dim light conditions, showing the paradox between the first- and second-order systems. Lin, Ma, et al. \cite{lin2022dvs} use Brownian motion with drift for simulation, but the simulated behaviors deviate partly from those of real DVS in dim light conditions. The above researches contribute greatly to the developments of DVS, but lack of further analysis of DVS in dim light conditions which are quite common in practical applications.  

In this paper, the typical DVS circuit is studied. We find that there exists discontinuity of event triggering time, which is one of the main factors deciding the DVS's unsatisfied behaviors in dim light conditions. Based on our analysis, the discontinuity extends with slow changing speed of light intensity. Because the dim light conditions generally share small difference of light intensity, the discontinuity will be prominent. By analyzing the properties of DVS circuit,  the discontinuity of event triggering time is attributed to the charge and discharge of parasitic capacitor of photodiode. Experimental results on real data of DVS validate the above analysis and will be helpful for further improvements of DVS, such as design of event simulators. For convenience, the meanings of symbols used in this paper are illustrated in Table. \ref{tbl:meaning table}.

\begin{table}
	\centering
	\caption{symbol meanings}
	\resizebox{\linewidth}{!}{
	\begin{tabular}{ |c|c| } 
		\hline
		symbols & meanings \\ 
		\hline
		$\rm M_{fb}$ & feedback MOS\\
		$\rm M_n$ and $\rm M_{cas}$ & cascode MOS amplifier\\ 
		$\rm M_{sf}$ & MOS source follower \\
		$\rm M_{dp}$ & MOS providing difference of voltage \\ 
		$\rm PD$ & photodiode\\
		$\rm C_1$ and $\rm C_2$ & capacitors for differential amplifier\\
		$\rm C_J$ & parasitic capacitor of PD \\
		$L$ & light intensity\\
		$I_{pd}$ & photocurrent across photodiode $\rm PD$ \\
		$V_{pd}$ & voltage of photodiode \\
		$V_{p}$ & gate voltage of $\rm M_{fb}$ fed back by cascode \\
		$V_{sf}$ & drain voltage of $\rm M_{sf}$ buffered from $V_p$ \\
		$V_d$ & source voltage of $\rm M_{dp}$\\ 
		$\Delta Q_e$ & charge of difference in $\rm C_J$ during triggering of events \\
		$\Delta t_e$ & time delay between consecutive triggering events \\
		$\mu$ & changing speed of light intensity \\
		\hline
	\end{tabular}
	}
	\label{tbl:meaning table}
\end{table}

\section{Discontinuity of event triggering time}
\label{sec:Discontinuity distribution of event triggering time}

In this section, the typical DVS circuit is analyzed along with its operation principle. The charge and discharge of parasitic capacitor of photodiode causes a time delay between triggering events, leading to the discontinuity. The non-first-order behavior is aroused by the duration of discontinuity that depends exclusively on the changing speed of  light intensity and photocurrent.

\subsection{DVS circuit and operation principle}
\label{subsec:DVS circuit and operation principle}

\begin{figure}[htb]
	\centering
	\includegraphics[width=8.5cm]{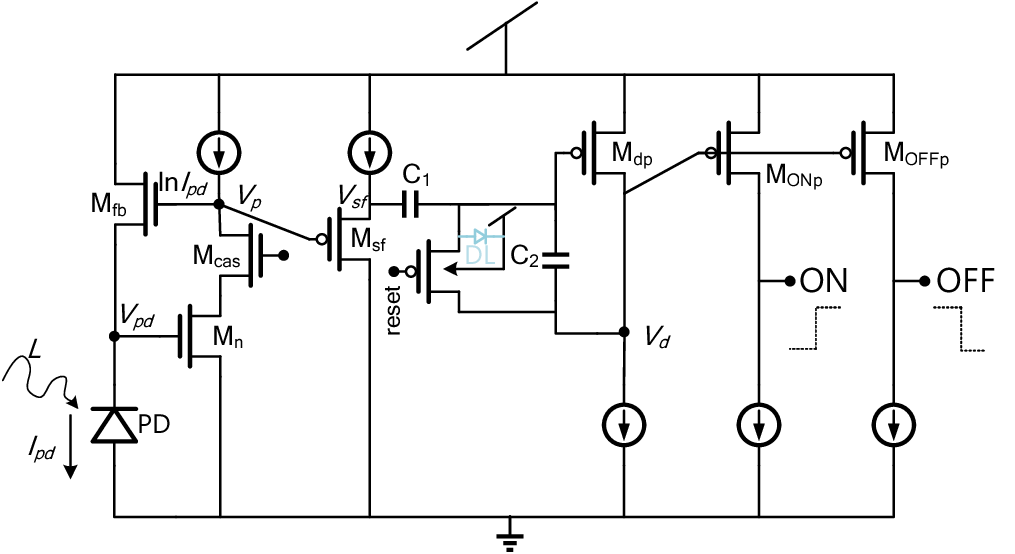}
	\caption{DVS pixel circuit.}
	\label{fig:pixel_circuit}
\end{figure}

The typical DVS circuit of one pixel is illustrated in ${\rm{Fig.} \ref{fig:pixel_circuit}}$. When the light signal on the photodiode $\rm PD$ is static without change, there will be few events other than some noise induced by the junction leak current of diode DL. Assume that the last event triggers at the time $t_1$.
After time $\Delta t$, the incident light on $\rm PD$ changes from $L(t_1)$ to ${L(t_1+\Delta t)}$, and the source voltage of the metal-oxide-semiconductor (MOS) transistor $\rm{M_{fb}}$ varies from \textit{$V_{d}(t_1)$} to ${V_{d}(t_1+\Delta t)}$. In case that the MOS comparators $(\rm{M_{ONp}} \quad and \quad \rm{M_{OFFp}})$ detect that ${V_{d}(t_1+\Delta t)}$ overcomes predefined ON threshold -$\Theta_{ON}$ or OFF threshold $\Theta_{OFF}$, an ON or OFF event is triggered, respectively. The process is formulated as follows:

\begin{equation}
	{V_{d}(t_1+\Delta t)} \quad \begin{cases}
	\geq \Theta_{OFF}, & trigger\ OFF\ events \\
	\leq -\Theta_{ON}, & trigger\ ON\ events \\
	others, & no\ events. \\
	\end{cases} 
	\label{threshold_detect}
\end{equation}
The source voltage ${V_{d}}$ will be reset when the event triggers. 

After the introduction of event triggering, we analyze the signal transduction from light signal $L(t)$ to electrical signal $V_{d}(t)$ in the DVS circuit. At time $t_1$, the photocurrent flowing across the photodiode $\rm PD$ is $I_{pd}(t_1)$. After time $\Delta t$, the change of photocurrent is induced in the photodiode $\rm PD$ that satisfies:
\begin{equation}
	I_{pd}(t_1 + \Delta t) - I_{pd}(t_1) \propto L(t_1 + \Delta t) - L(t_1)
	\label{Ipd and L}.  
\end{equation}

The differential amplifier with capacitors $\rm C_1$ and $\rm C_2$ amplifies $V_{sf}$ that is buffered from $V_p$ and then drives the gate of MOS $\rm M_{dp}$ to control the source voltage $V_d$ \cite{delbruck2008IEEEJSSC}: 
\begin{eqnarray}    \label{delta_Vd}
	\Delta V_{d}(t_1)&=&-A \cdot \Delta V_{sf}\nonumber    \\
	&=&\frac{-A \cdot V_T \cdot \kappa_{sf}}{\kappa_{fb}}({\rm ln}I_{pd}(t_1 + \Delta t) - {\rm ln}I_{pd}(t_1)), \nonumber    \\
\end{eqnarray}
where $A=\frac{\rm C_1}{\rm C_2}$ is the gain of differential amplifier,  $\kappa_{sf}$ and $\kappa_{fb}$ are the slope factors of MOS $\rm M_{sf}$ and  $\rm M_{fb}$, respectively, and $V_T$ is the thermal voltage.

\subsection{parasitic capacitor of photodiode} 

Actually, the photodiode voltage $V_{pd}$ changes with the light signal $L$, despite the clamping by the feedback loop and cascode. The cascode of $\rm M_n$ and $\rm M_{cas}$ senses the change of $V_{pd}$ with amplification. The amplified result $V_p$ is fed back by $\rm M_{fb}$ to accommodate the change of $I_{pd}$.  

\begin{figure}[htb]
	\centering
	\includegraphics[width=8.5cm]{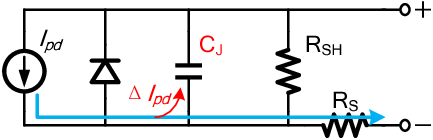}
	\caption{distributed diagram of photodiode PD.}
	\label{fig:distribute_photodiode}
\end{figure}

In the distributed diagram of photodiode $\rm PD$ given in ${\rm{Fig.} \ref{fig:distribute_photodiode}}$, the junction of $\rm PD$ consists of the parasitic capacitor $\rm C_J$. As the parallel resistance $\rm R_{SH}$ is extremely high and the series resistance $\rm R_{S}$ tends to be quite low \cite{dentan1990numerical}, the charge and discharge of $\rm C_J$ is one of the main reasons for the delay and discontinuity of event triggering time.

Ideally, when there is a light intensity change in time $\Delta t$, an immediately stimulated current $\Delta I_{pd} = I_{pd}(t_1+\Delta t) - I_{pd}(t_1)$ will flow across $\rm M_{fb}$. However, in fact this is not the case. In the distributed diagram of photodiode $\rm PD$ shown in ${\rm{Fig.} \ref{fig:distribute_photodiode}}$, $\Delta I_{pd}$ needs to charge or discharge the parasitic capacitor $\rm C_J$  of $\rm PD$, firstly. The process is not completed until the voltage of $\rm C_J$ reaches $V_{pd}(t_1+\Delta t)$ to accommodate the stimulated current $\Delta I_{pd}$. 

\subsection{time delay and discontinuity} 

The charge and discharge of parasitic capacitor $\rm C_J$ is time-consuming, leading to a time delay between triggering events. If we evaluate the triggering time of events, the distribution of triggering time is probably discontinuous because there are few events in the time delay region.  

As the photodiode $\rm PD$ is reverse-biased, the barrier capacitance is dominant in $\rm C_J$. Compared with the built-in potential of photodiode $\rm PD$ (usually hundreds of mV) \cite{razavi2021fundamentals}, the voltage change of $V_{pd}$ is much smaller (nearly a few mV) \cite{gracca2023shining}. According to \cite{razavi2021fundamentals}, the parasitic capacitance of $\rm C_J$ is almost unchanged.

The change of electric charge $\Delta Q$ stored in the capacitor $\rm C_J$ is:
\begin{equation}
	\Delta Q=\Delta V_{pd} \cdot {\rm C_J} = \Delta I_{pd} \cdot \Delta t
	\label{charge relation}.  
\end{equation}
 
Based on Eqs. \ref{threshold_detect} and \ref{delta_Vd}, the change $\Delta V_p$ is fixed between two consecutive events. In the small-signal analysis of cascode,

\begin{equation}
	\Delta V_{p}=-A_{cas} \cdot \Delta V_{pd}
	\label{cascode vpd},  
\end{equation}
where $-A_{cas}$ is the small-signal gain of cascode.

Therefore, $\Delta V_{pd}$ and $\Delta Q$ are invariable between two consecutive triggering events if we do not consider the influence by shot noise, dark current, and leak noise. Assuming that $\Delta Q_e$ is the fixed electric charge difference, there is a time delay between triggering events:

\begin{equation}
	\Delta t_{e}=\frac{\Delta Q_{e}}{\Delta I_{pd}}
	\label{time_delay}.  
\end{equation} 

Note that: 1. $\Delta t_{e}$ is not proportional to $\frac{\Delta L}{L}$ ($\Delta \ln  L$), which means that the time delay follows a non-first-order system rule of DVS; 2. As {$\Delta I_{pd}$} is only related to the threshold ($\Theta_{ON}$ and $\Theta_{OFF}$),\ $\Delta t_{e}$ is invariant to current light intensity. 

In fact, {$\Delta I_{pd}$} needs some time to change because light intensity varies with time. DVS voltmeter \cite{lin2022dvs} supposes that {$\Delta I_{pd}$} and {$\Delta L$} change linearly with time, which is a reasonable assumption in a short period. In that case, $\Delta t_{e}$ follows: 

\begin{equation}
	\Delta t_e \propto \frac{1}{\Delta I_{pd}} \propto \frac{1}{\mu \cdot T}
	\label{mu},  
\end{equation}
where $\mu=\frac{L(t_1+\Delta t) - L(t_1)}{\Delta t}$ is the changing speed of light intensity and $T$ is the total time between the two triggering events. As can be seen, the time delay $\Delta t_e$ is inversely proportion to the changing speed of light intensity $\mu$.

The existence of $\Delta t_e$ delays the triggering of events, breaking the inverse Gaussian distribution that is followed by the event triggering time \cite{lin2022dvs}. Because that there are few events triggering during the time delay $\Delta t_{e}$, the distribution of event triggering time fluctuates dramatically, appearing as the discontinuity in the histograms and probability density functions (PDF). As a result, the length of discontinuity in the histograms and PDFs of event triggering time is $\Delta t_{e}$. 

 In the dim light conditions, the difference of light intensity is generally small, leading to a slow changing speed of light. Therefore, the time delay $\Delta t_e$ is prominent.

\section{EXPERIMENTS}
\label{sec:experiments}

In the section, we provide the existence of time delay $\Delta t_{e}$ and the discontinuity of event triggering time based on real experimental data. The details of time delay $\Delta t_{e}$ are further validated.

\subsection{experimental datasets and settings}
A real public DAVIS dataset \cite{mueggler2017event}, captured by DAVIS240 event camera, is used. Inside the dataset, \textit{boxes\underline{\hspace{0.5em}}6dof} scene including over 1.3 billion events is applied. To evaluate the time delay and discontinuity of event triggering time accurately, the frame-based videos are interpolated \cite{jiang2018super} by 10 times, as done in \cite{lin2022dvs, hu2021v2e,rebecq2018esim}. We use the ITU-R recommendation $BT.\ 709$ for linear conversion \cite{series2017colour}. For the gray frame-based videos, the pixel value is treated as the luma value, in the changing speed $\mu$ and light intensity $L$. 

In the following experiments, the histograms of event triggering time are collected under different light intensity $L$ and light changing speed $\mu$. Here $L$ is specified as the average pixel value between the two consecutive events. 

\subsection{results}

\begin{figure*}
	\centering
	\begin{subfigure}[b]{0.19\textwidth}
		\centering
		\includegraphics[width=\textwidth]{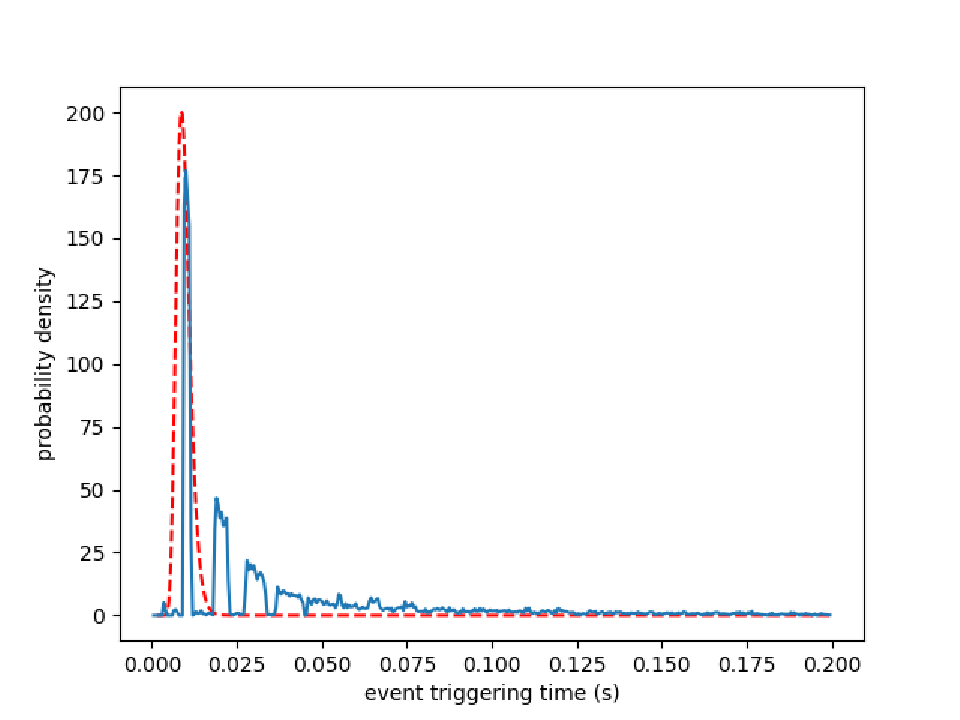}
		\caption{$\mu=50,\ L=10$}
		\label{dl50al10}
	\end{subfigure}
	\hspace{0.1mm}
	\begin{subfigure}[b]{0.19\textwidth}
		\centering
		\includegraphics[width=\textwidth]{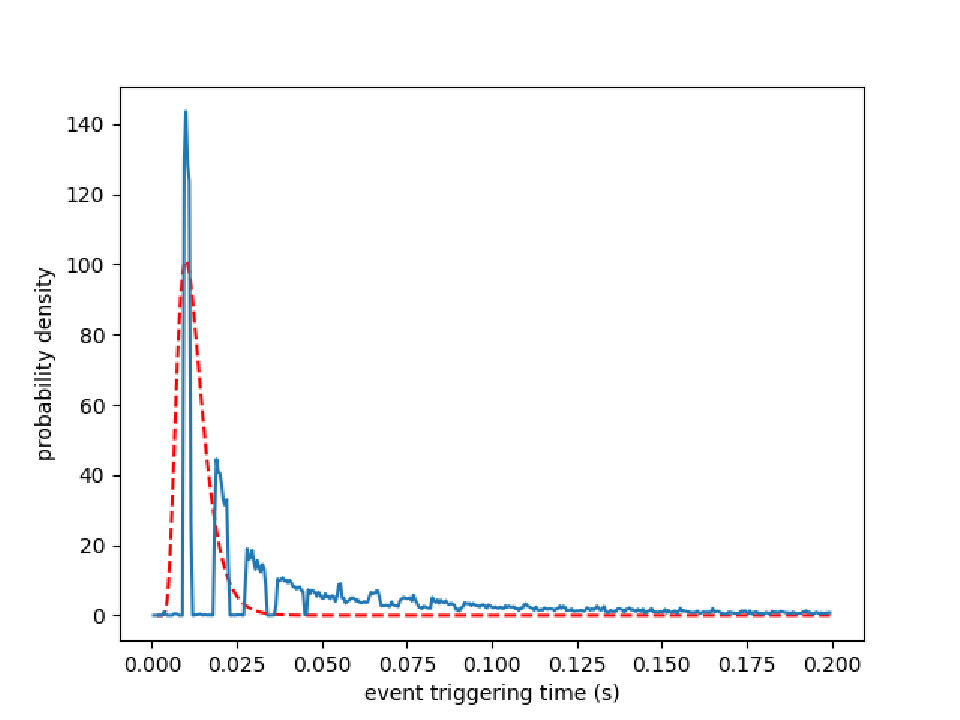}
		\caption{$\mu=50,\ L=20$}
		\label{dl50al20}
	\end{subfigure}
	\hspace{0.1mm}
	\begin{subfigure}[b]{0.19\textwidth}
		\centering
		\includegraphics[width=\textwidth]{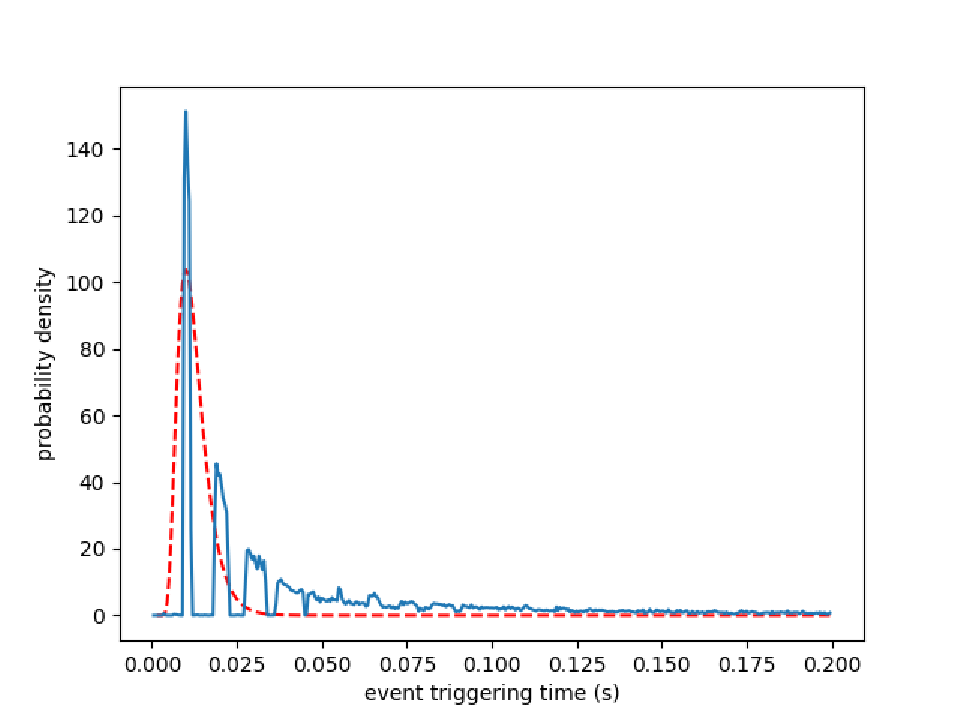}
		\caption{$\mu=50,\ L=30$}
		\label{dl50al30}
	\end{subfigure}
	\hspace{0.1mm}
	\begin{subfigure}[b]{0.19\textwidth}
		\centering
		\includegraphics[width=\textwidth]{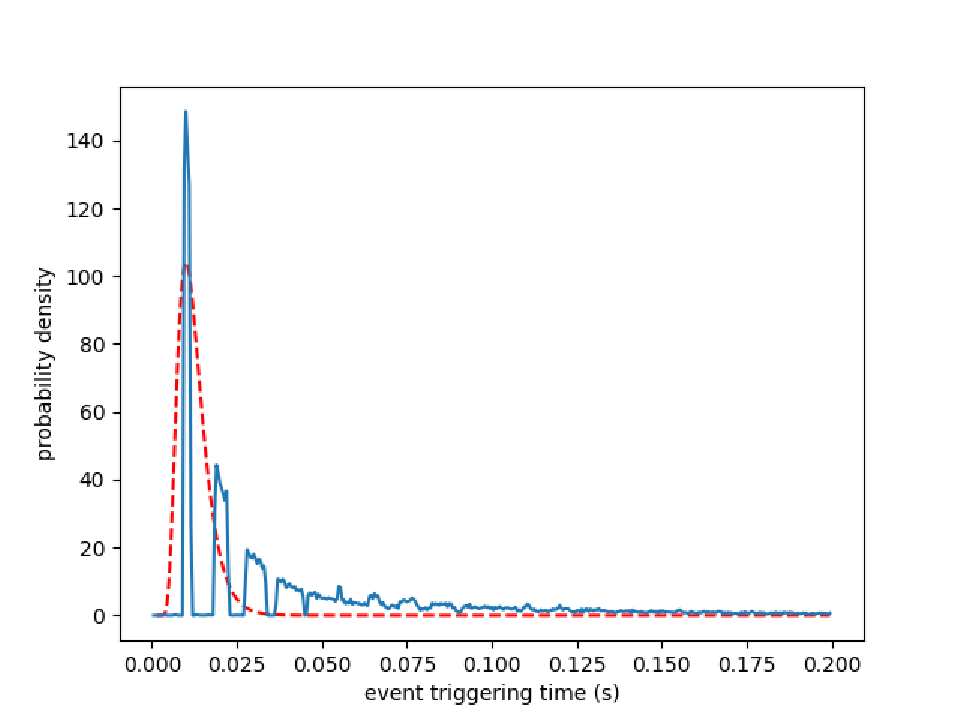}
		\caption{$\mu=50,\ L=40$}
		\label{dl50al40}
	\end{subfigure}
	\hspace{0.1mm}
	\begin{subfigure}[b]{0.19\textwidth}
		\centering
		\includegraphics[width=\textwidth]{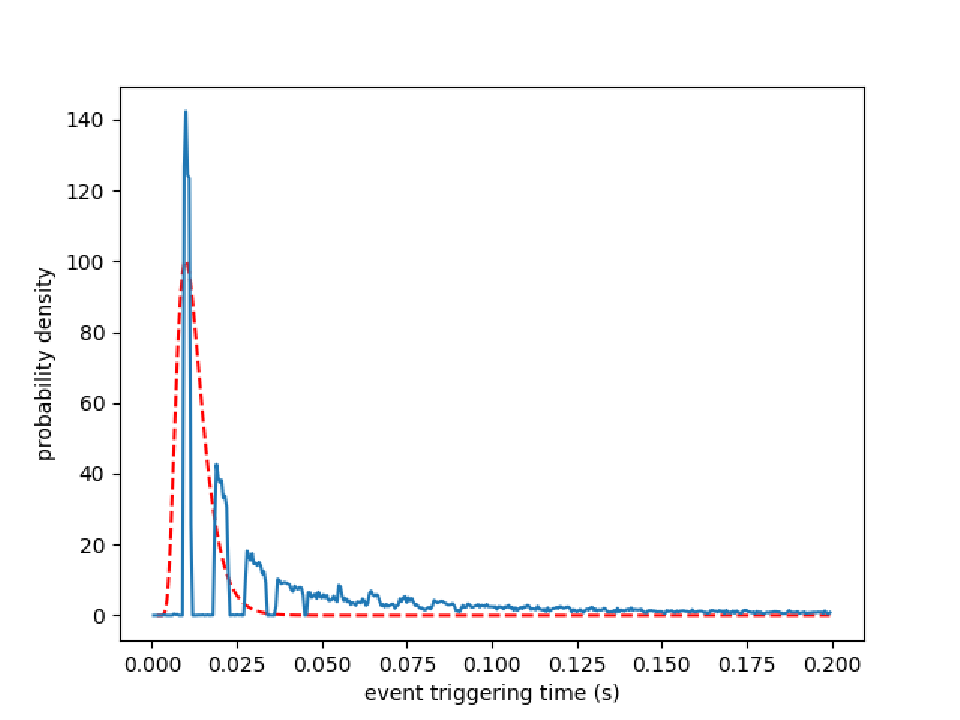}
		\caption{$\mu=50,\ L=50$}
		\label{dl50al50}
	\end{subfigure}
	
	\begin{subfigure}[b]{0.19\textwidth}
		\centering
		\includegraphics[width=\textwidth]{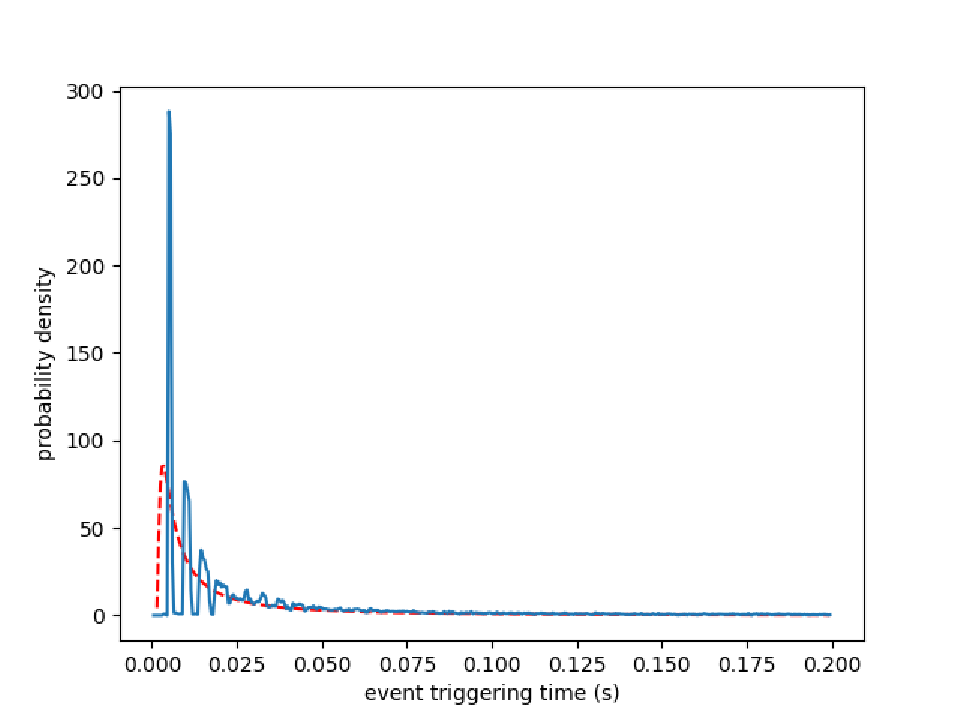}
		\caption{$\mu=100,\ L=10$}
		\label{dl100al10}
	\end{subfigure}
	\hspace{0.1mm}
	\begin{subfigure}[b]{0.19\textwidth}
		\centering
		\includegraphics[width=\textwidth]{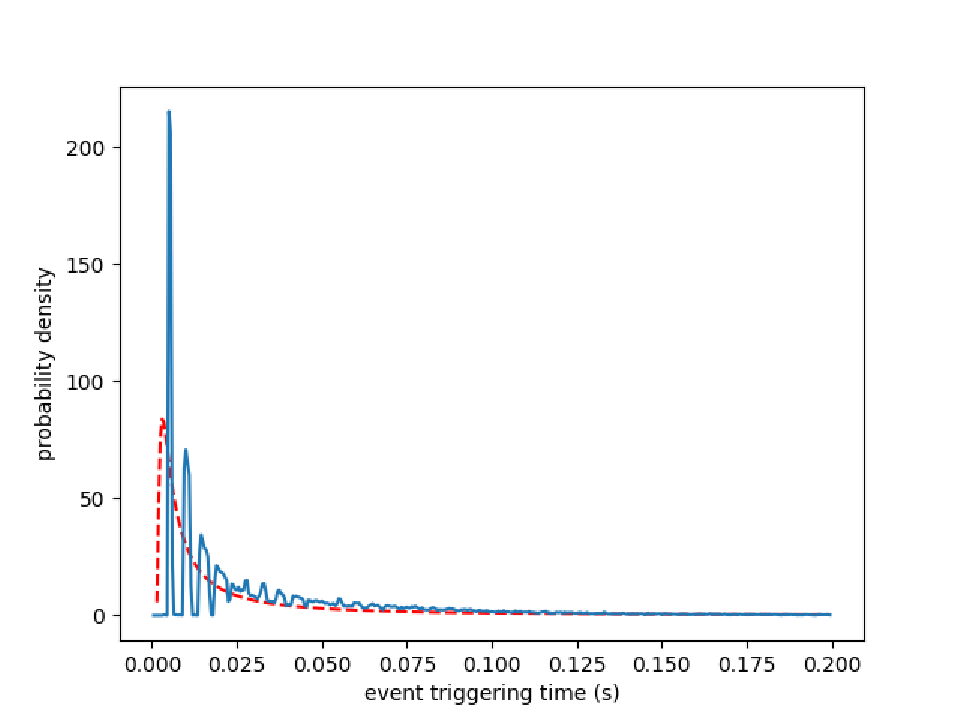}
		\caption{$\mu=100,\ L=20$}
		\label{dl100al20}
	\end{subfigure}
	\hspace{0.1mm}
	\begin{subfigure}[b]{0.19\textwidth}
		\centering
		\includegraphics[width=\textwidth]{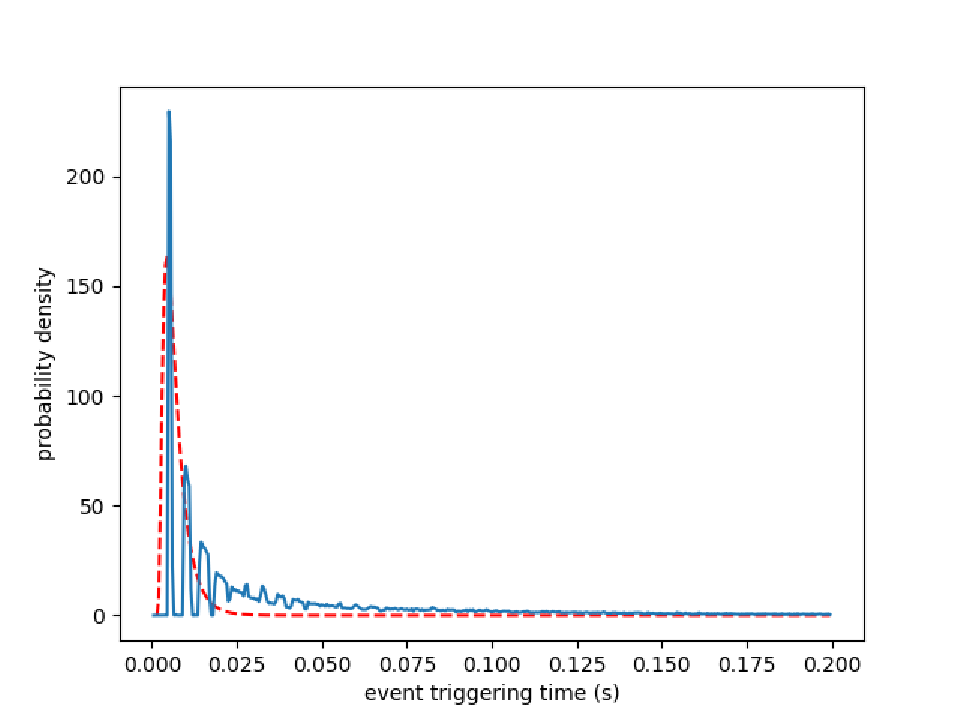}
		\caption{$\mu=100,\ L=30$}
		\label{dl100al30}
	\end{subfigure}
	\hspace{2mm}
	\begin{subfigure}[b]{0.19\textwidth}
		\centering
		\includegraphics[width=\textwidth]{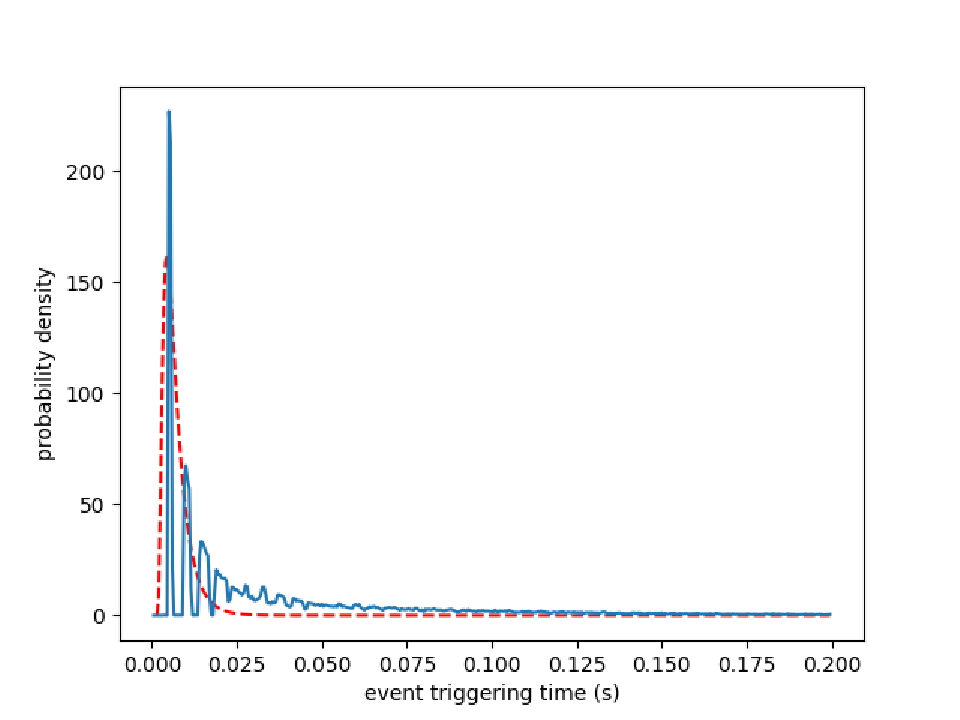}
		\caption{$\mu=100,\ L=40$}
		\label{dl100al40}
	\end{subfigure}
	\hspace{0.1mm}
	\begin{subfigure}[b]{0.19\textwidth}
		\centering
		\includegraphics[width=\textwidth]{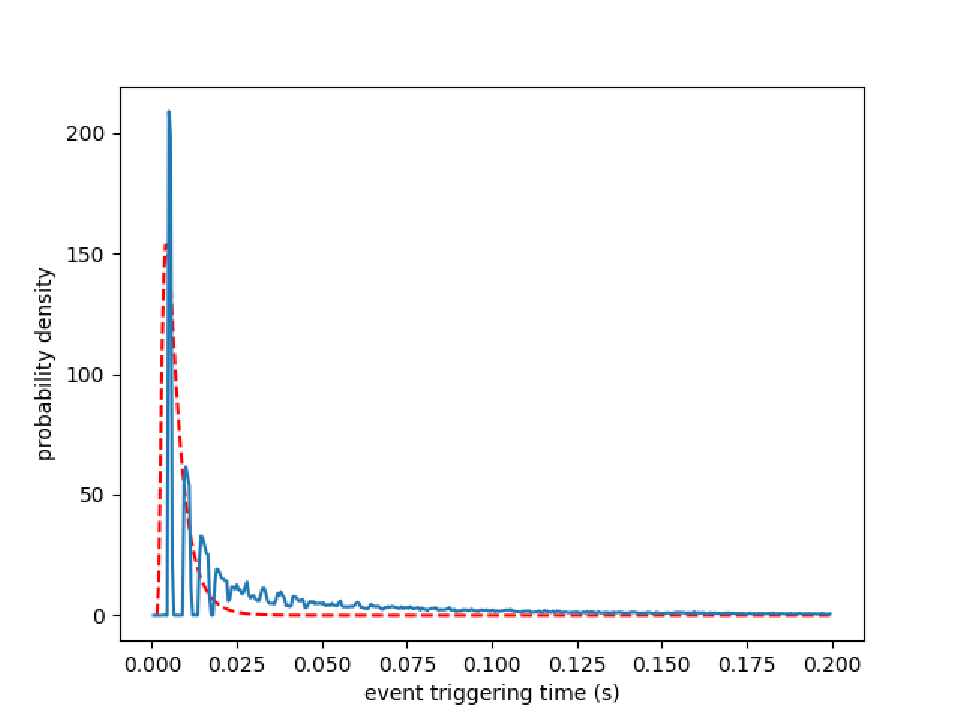}
		\caption{$\mu=100,\ L=50$}
		\label{dl100al50}
	\end{subfigure}
	
	\begin{subfigure}[b]{0.19\textwidth}
		\centering
		\includegraphics[width=\textwidth]{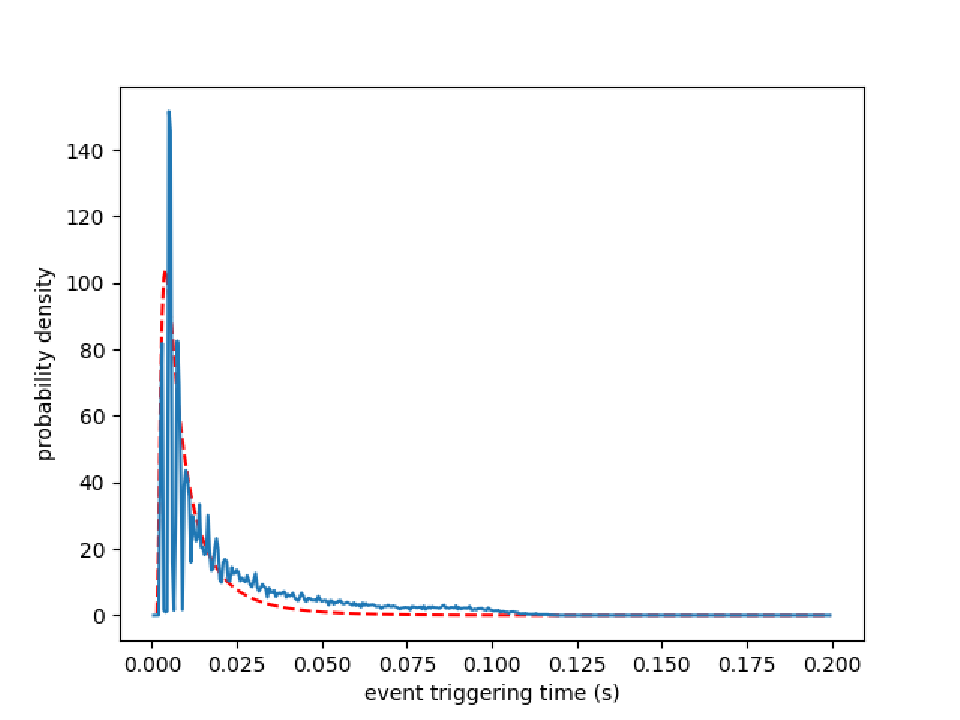}
		\caption{$\mu=200,\ L=10$}
		\label{dl200al10}
	\end{subfigure}
	\hspace{0.1mm}
	\begin{subfigure}[b]{0.19\textwidth}
		\centering
		\includegraphics[width=\textwidth]{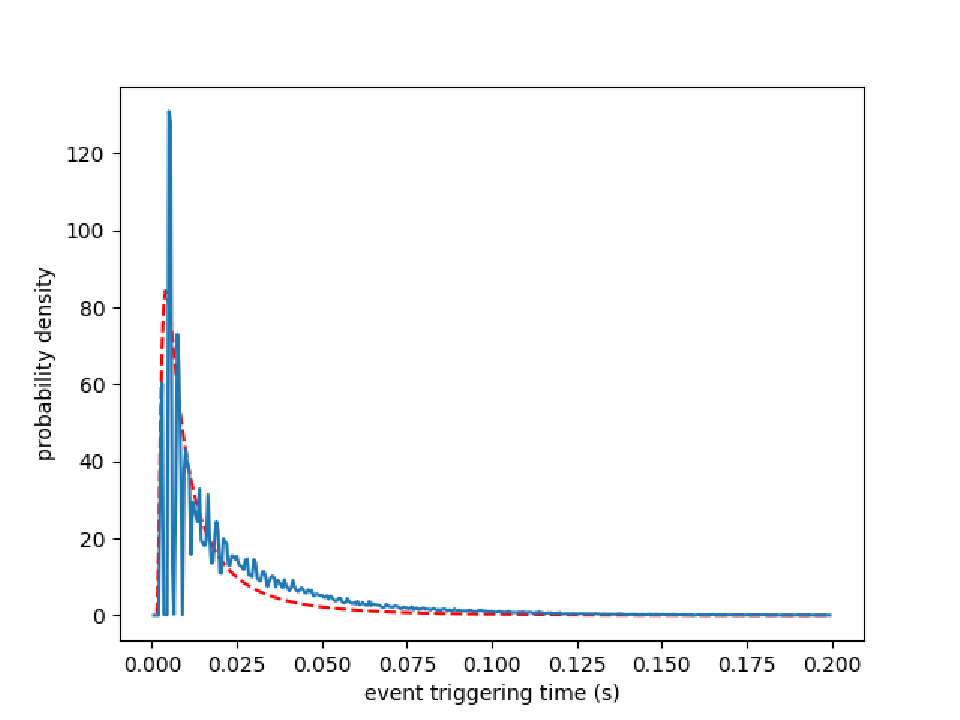}
		\caption{$\mu=200,\ L=20$}
		\label{dl200al20}
	\end{subfigure}
	\hspace{0.1mm}
	\begin{subfigure}[b]{0.19\textwidth}
		\centering
		\includegraphics[width=\textwidth]{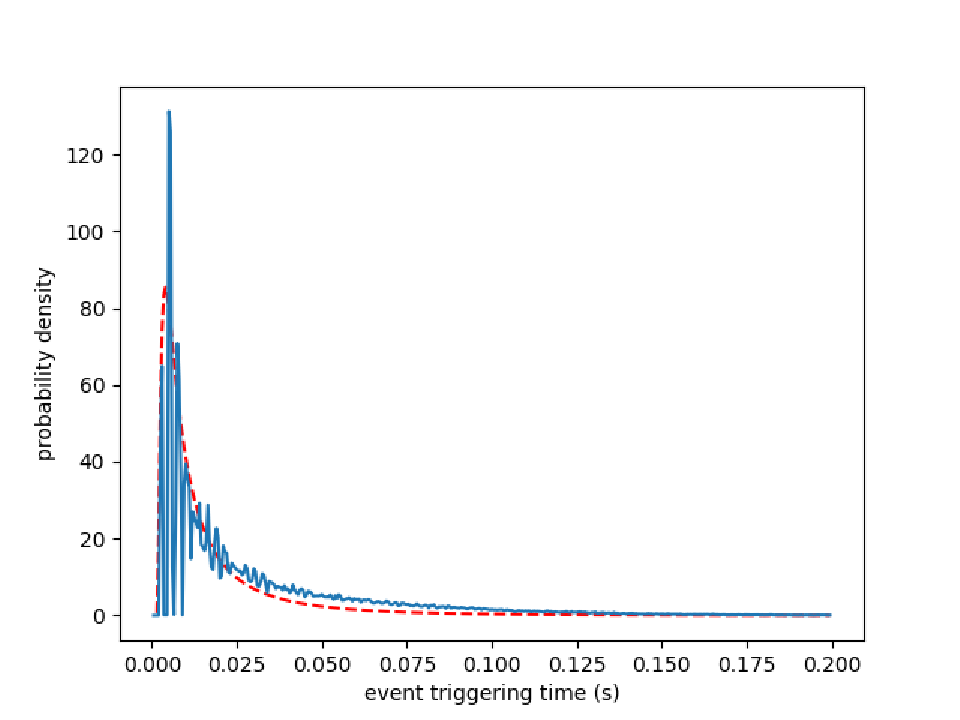}
		\caption{$\mu=200,\ L=30$}
		\label{dl200al30}
	\end{subfigure}
	\hspace{2mm}
	\begin{subfigure}[b]{0.19\textwidth}
		\centering
		\includegraphics[width=\textwidth]{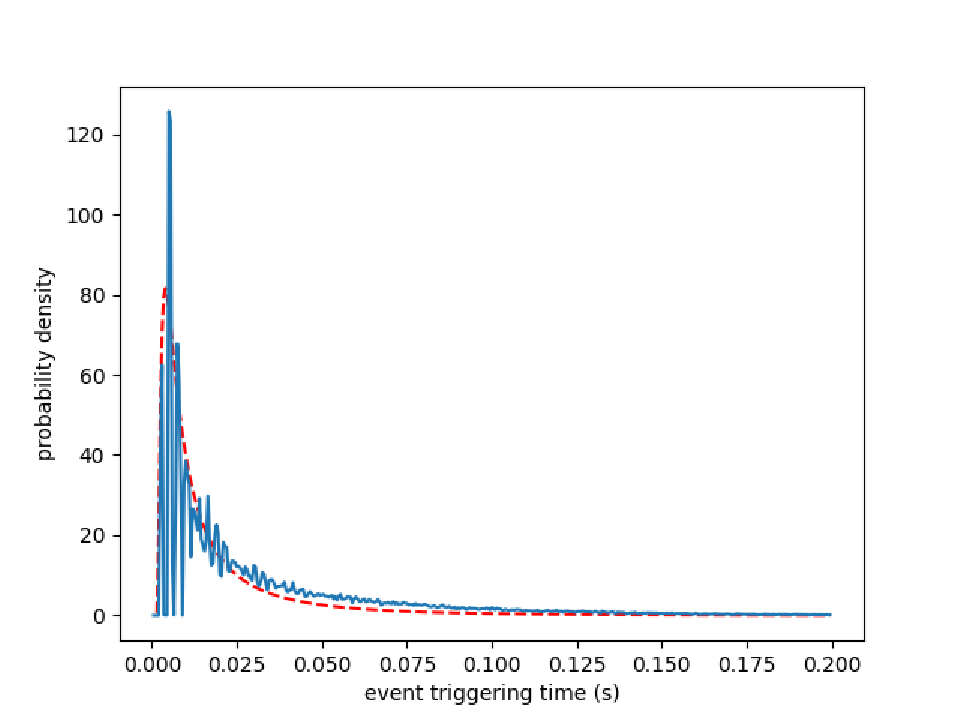}
		\caption{$\mu=200,\ L=40$}
		\label{dl200al40}
	\end{subfigure}
	\hspace{0.1mm}
	\begin{subfigure}[b]{0.19\textwidth}
		\centering
		\includegraphics[width=\textwidth]{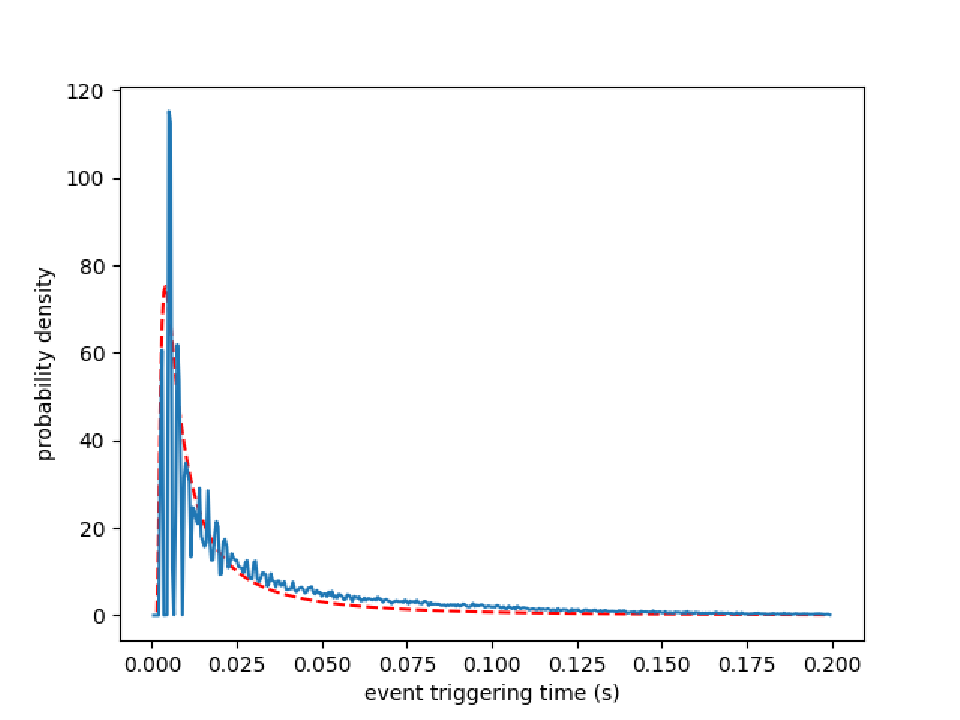}
		\caption{$\mu=200,\ L=50$}
		\label{dl200al50}
	\end{subfigure}
	
	\begin{subfigure}[b]{0.19\textwidth}
		\centering
		\includegraphics[width=\textwidth]{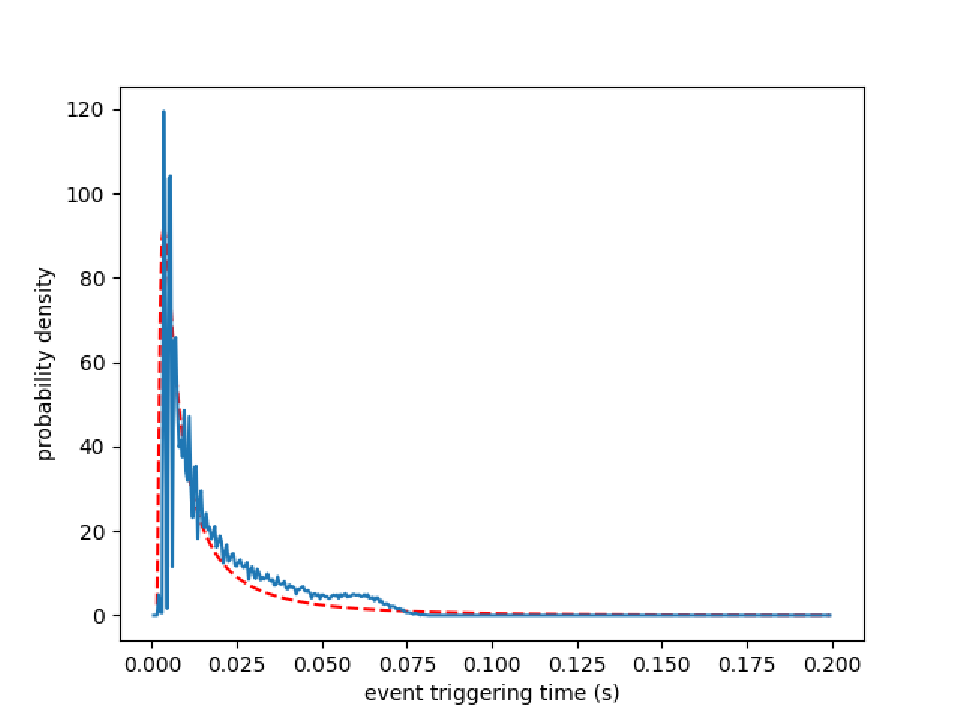}
		\caption{$\mu=300,\ L=10$}
		\label{dl300al10}
	\end{subfigure}
	\hspace{0.1mm}
	\begin{subfigure}[b]{0.19\textwidth}
		\centering
		\includegraphics[width=\textwidth]{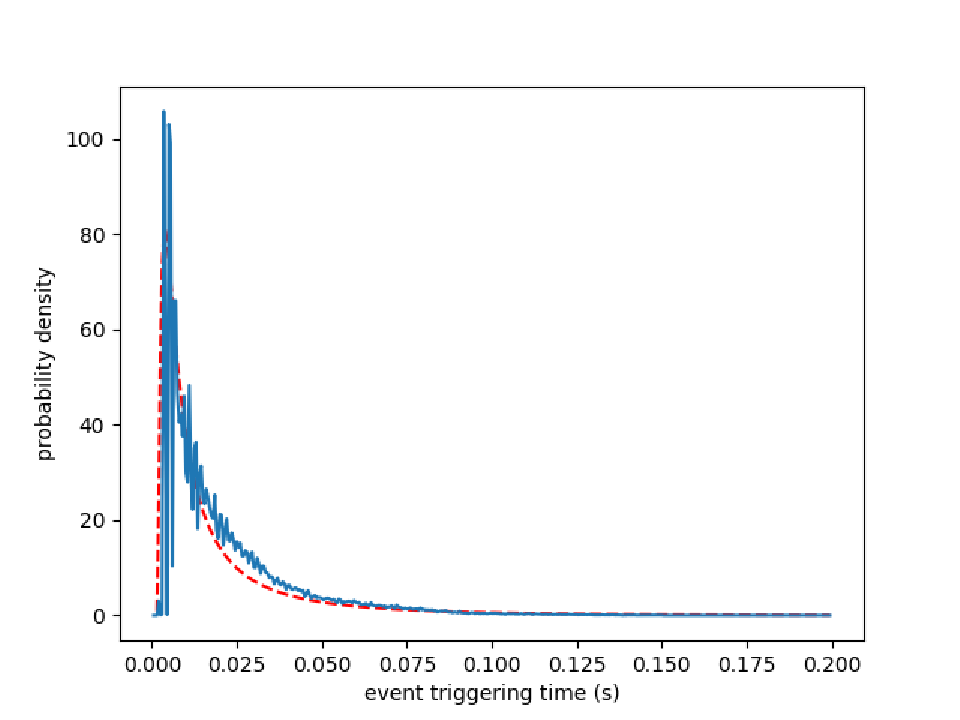}
		\caption{$\mu=300,\ L=20$}
		\label{dl300al20}
	\end{subfigure}
	\hspace{0.1mm}
	\begin{subfigure}[b]{0.19\textwidth}
		\centering
		\includegraphics[width=\textwidth]{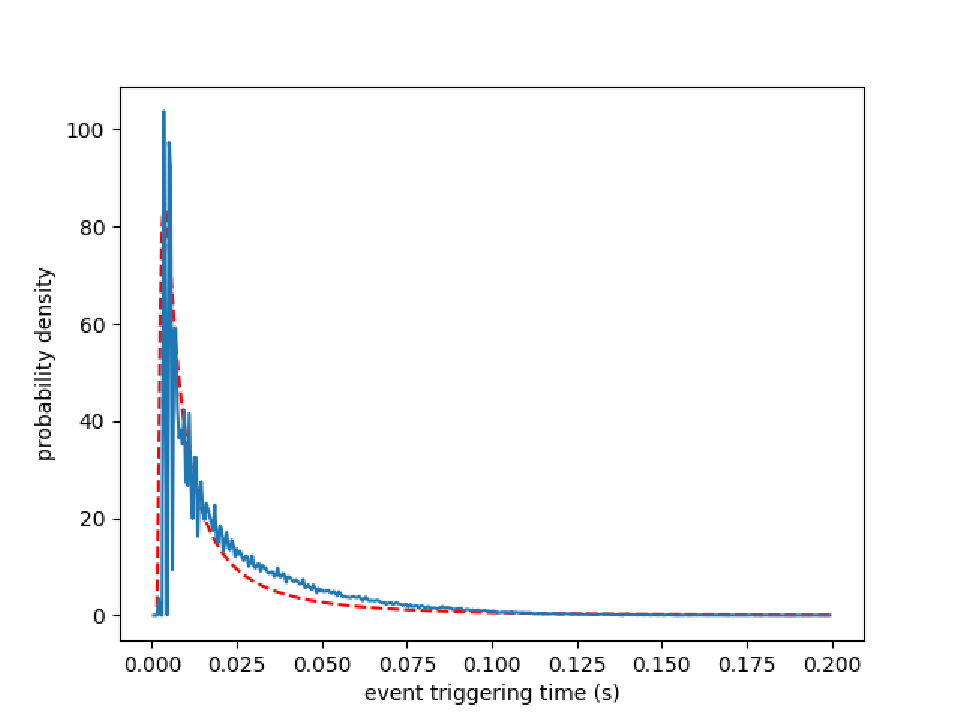}
		\caption{$\mu=300,\ L=30$}
		\label{dl300al30}
	\end{subfigure}
	\hspace{2mm}
	\begin{subfigure}[b]{0.19\textwidth}
		\centering
		\includegraphics[width=\textwidth]{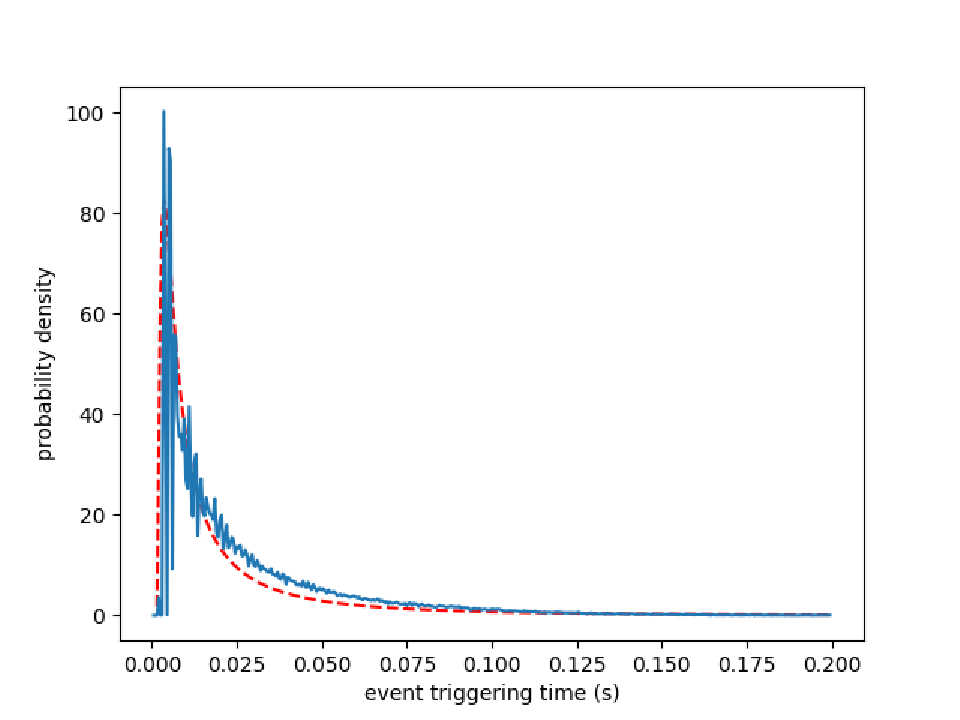}
		\caption{$\mu=300,\ L=40$}
		\label{dl300al40}
	\end{subfigure}
	\hspace{0.1mm}
	\begin{subfigure}[b]{0.19\textwidth}
		\centering
		\includegraphics[width=\textwidth]{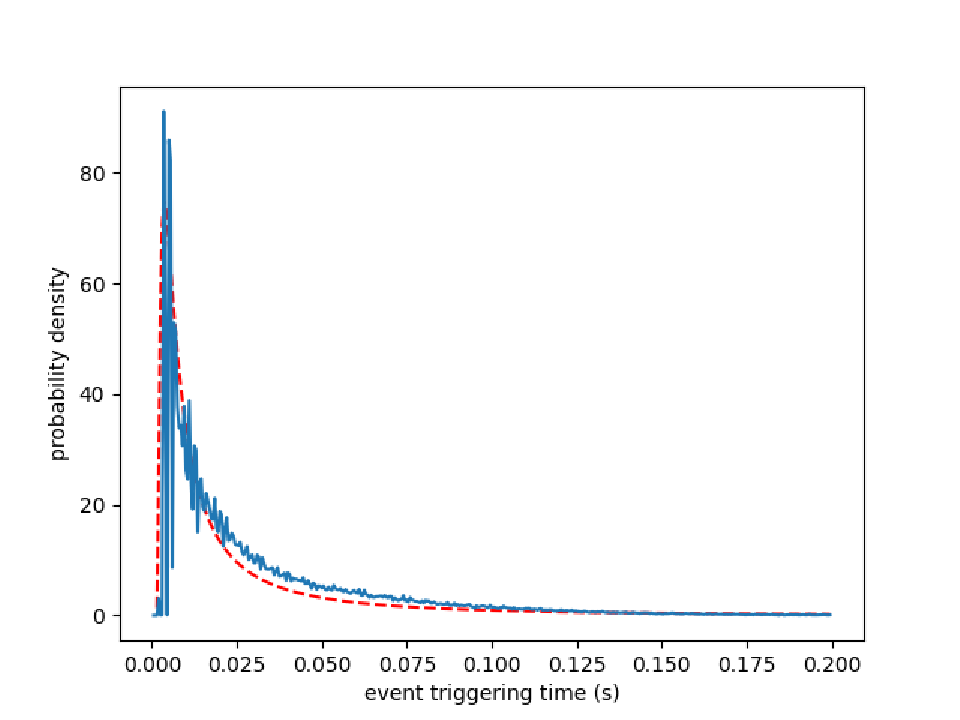}
		\caption{$\mu=300,\ L=50$}
		\label{dl300al50}
	\end{subfigure}
	
	\begin{subfigure}[b]{0.19\textwidth}
		\centering
		\includegraphics[width=\textwidth]{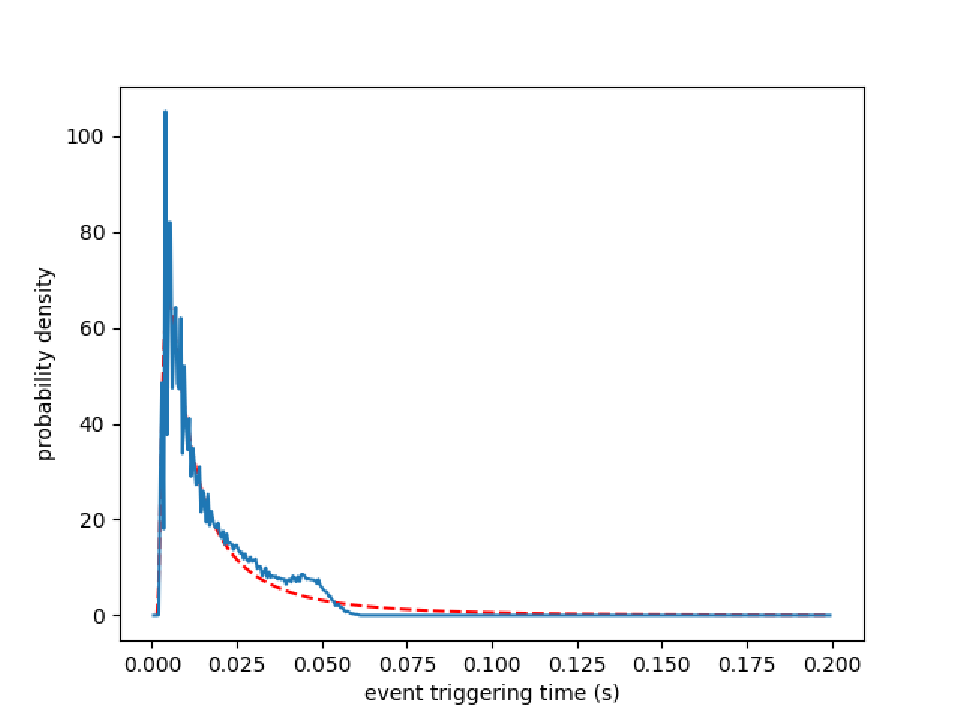}
		\caption{$\mu=400,\ L=10$}
		\label{dl400al10}
	\end{subfigure}
	\hspace{0.1mm}
	\begin{subfigure}[b]{0.19\textwidth}
		\centering
		\includegraphics[width=\textwidth]{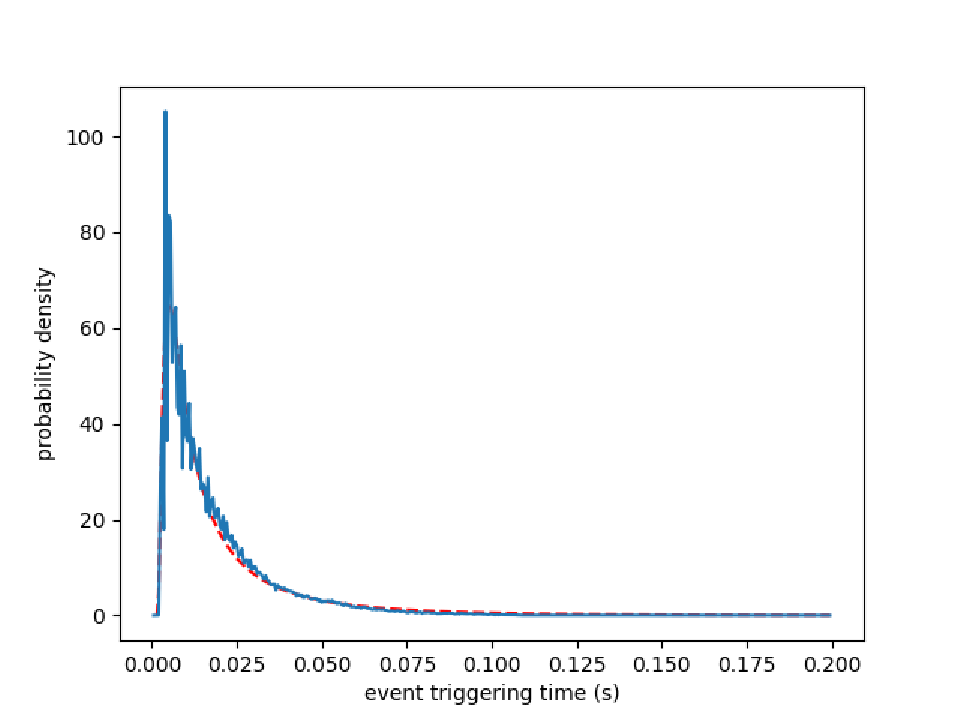}
		\caption{$\mu=400,\ L=20$}
		\label{dl400al20}
	\end{subfigure}
	\hspace{0.1mm}
	\begin{subfigure}[b]{0.19\textwidth}
		\centering
		\includegraphics[width=\textwidth]{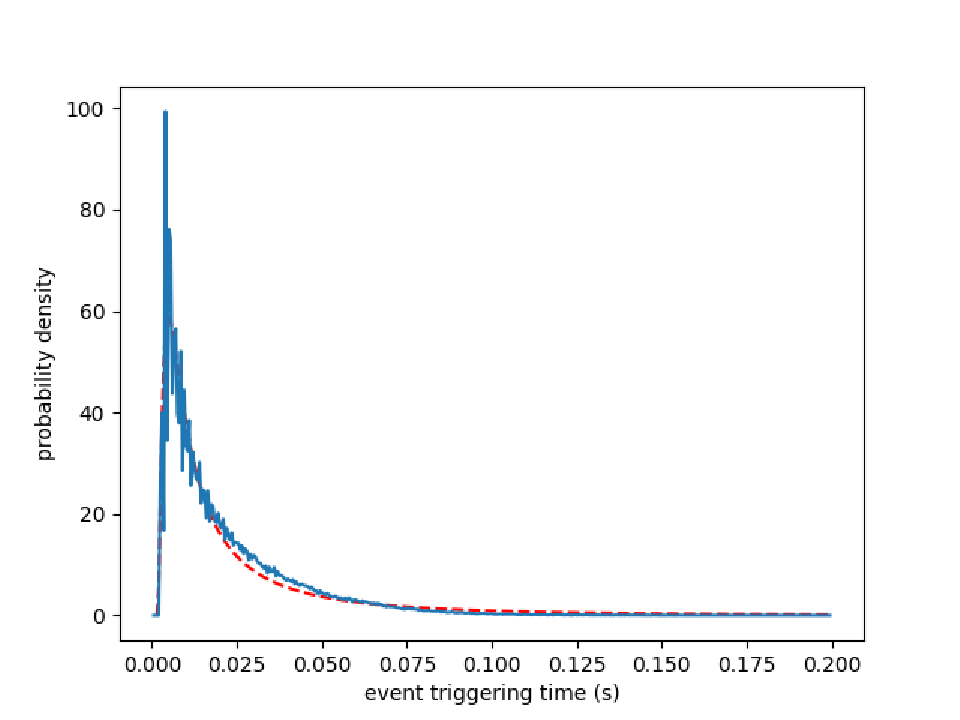}
		\caption{$\mu=400,\ L=30$}
		\label{dl400al30}
	\end{subfigure}
	\hspace{2mm}
	\begin{subfigure}[b]{0.19\textwidth}
		\centering
		\includegraphics[width=\textwidth]{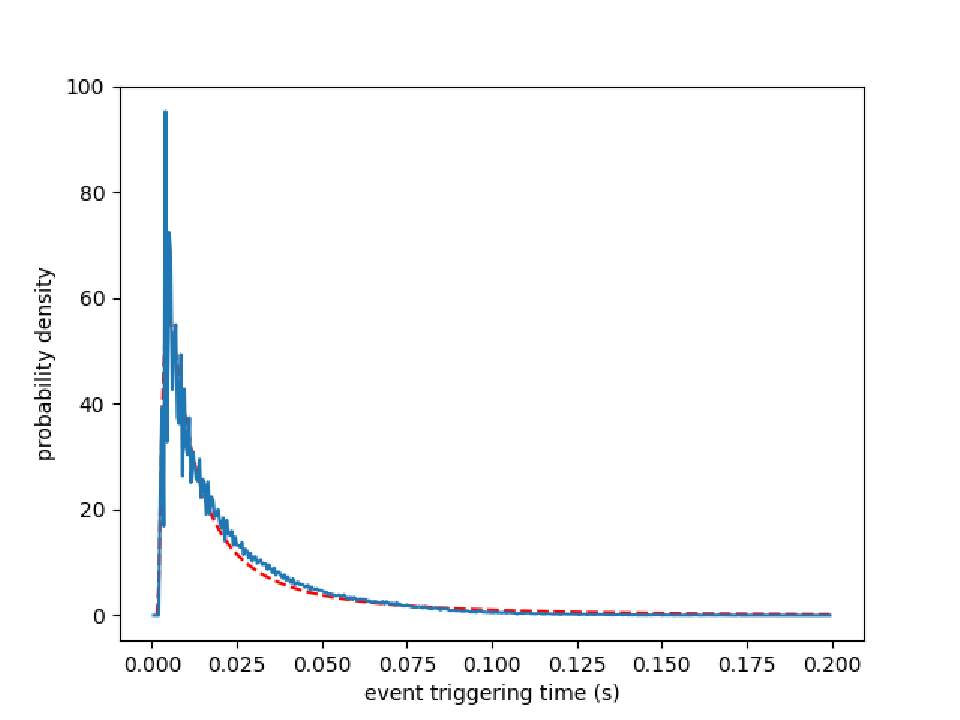}
		\caption{$\mu=400,\ L=40$}
		\label{dl400al40}
	\end{subfigure}
	\hspace{0.1mm}
	\begin{subfigure}[b]{0.19\textwidth}
		\centering
		\includegraphics[width=\textwidth]{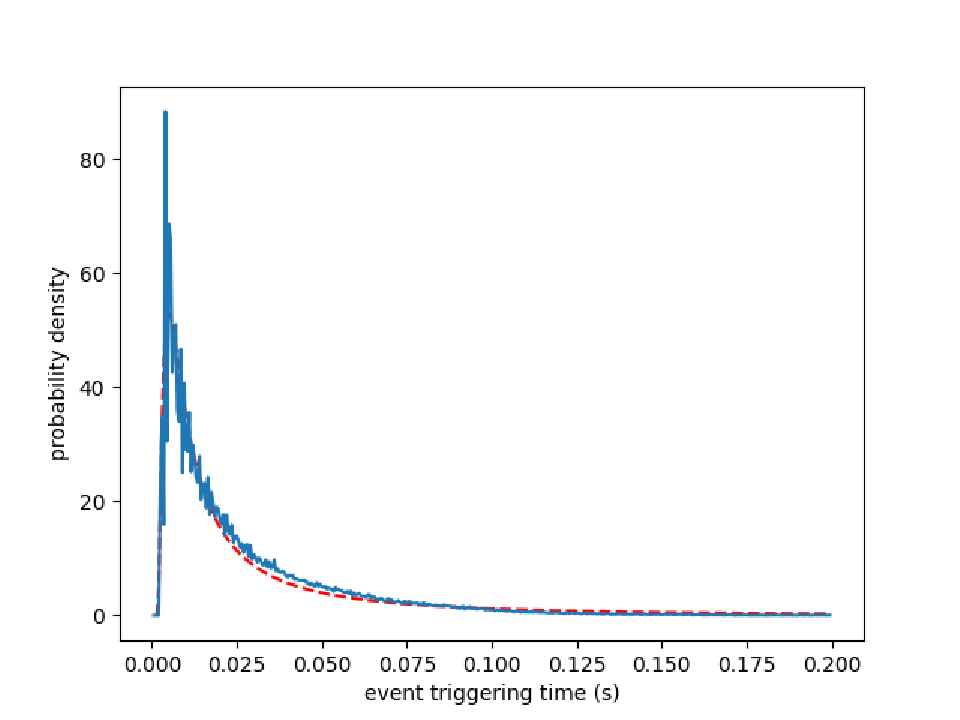}
		\caption{$\mu=400,\ L=50$}
		\label{dl400al50}
	\end{subfigure}
	
	\begin{subfigure}[b]{0.19\textwidth}
		\centering
		\includegraphics[width=\textwidth]{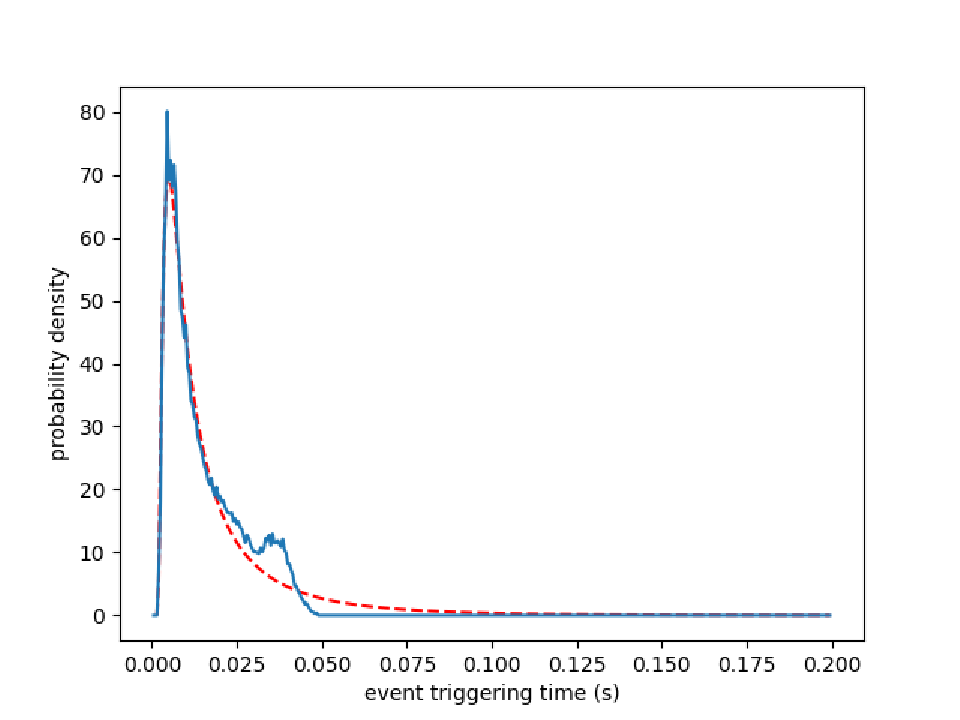}
		\caption{$\mu=500,\ L=10$}
		\label{dl500al10}
	\end{subfigure}
	\hspace{0.1mm}
	\begin{subfigure}[b]{0.19\textwidth}
		\centering
		\includegraphics[width=\textwidth]{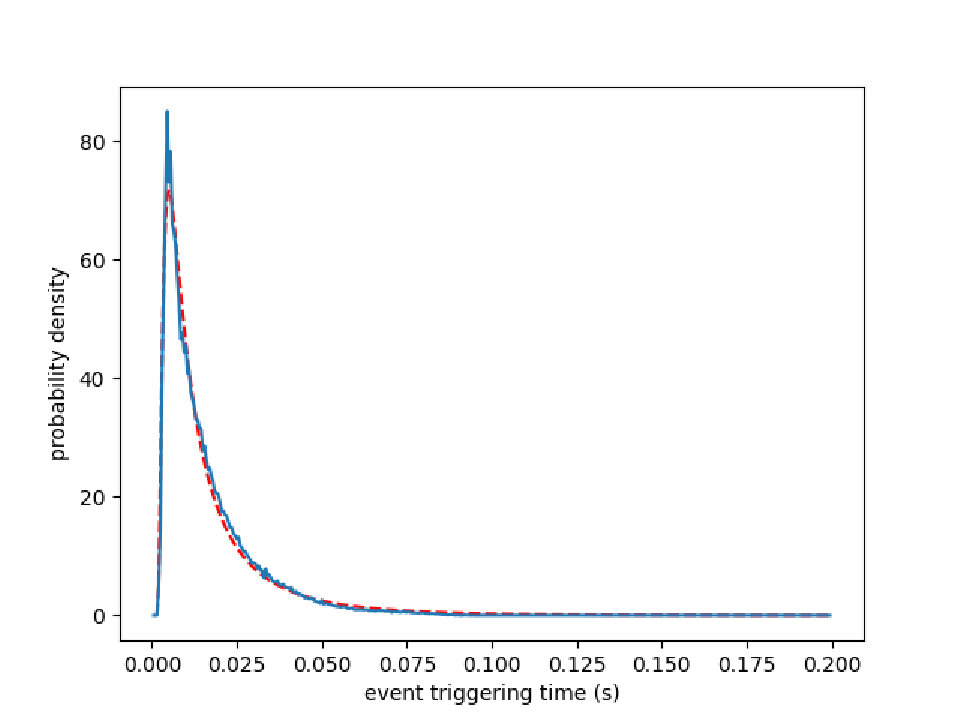}
		\caption{$\mu=500,\ L=20$}
		\label{dl500al20}
	\end{subfigure}
	\hspace{0.1mm}
	\begin{subfigure}[b]{0.19\textwidth}
		\centering
		\includegraphics[width=\textwidth]{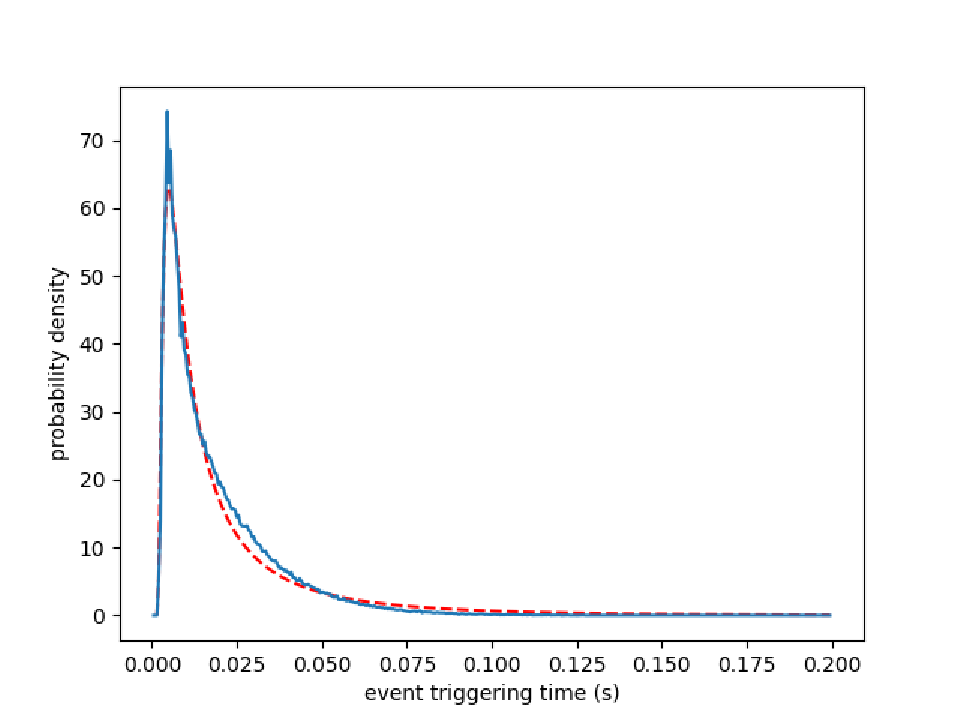}
		\caption{$\mu=500,\ L=30$}
		\label{dl500al30}
	\end{subfigure}
	\hspace{2mm}
	\begin{subfigure}[b]{0.19\textwidth}
		\centering
		\includegraphics[width=\textwidth]{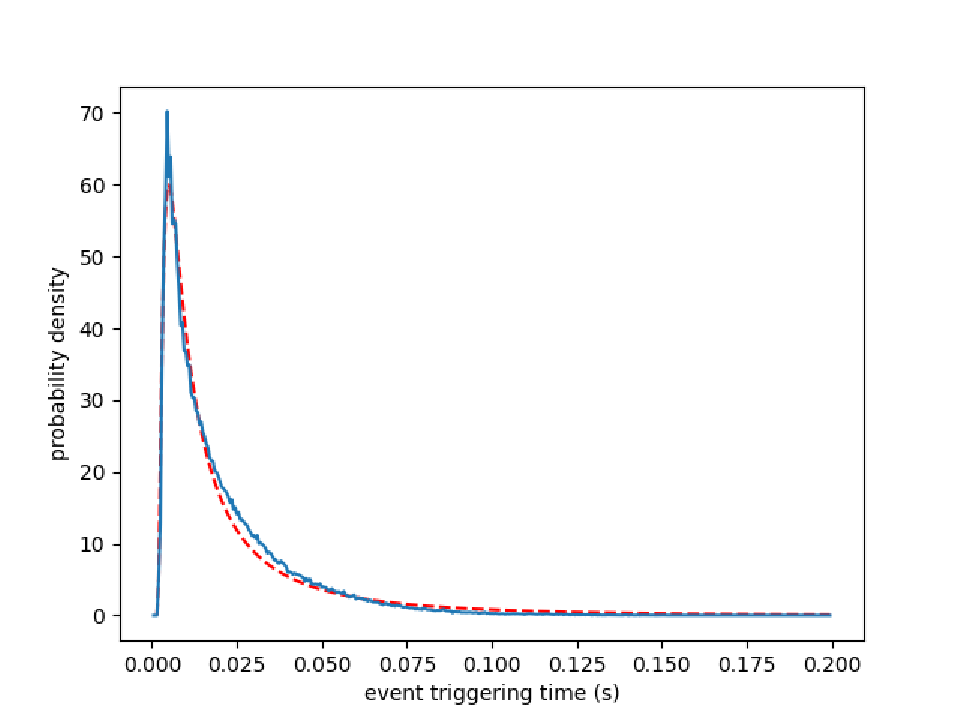}
		\caption{$\mu=500,\ L=40$}
		\label{dl500al40}
	\end{subfigure}
	\hspace{0.1mm}
	\begin{subfigure}[b]{0.19\textwidth}
		\centering
		\includegraphics[width=\textwidth]{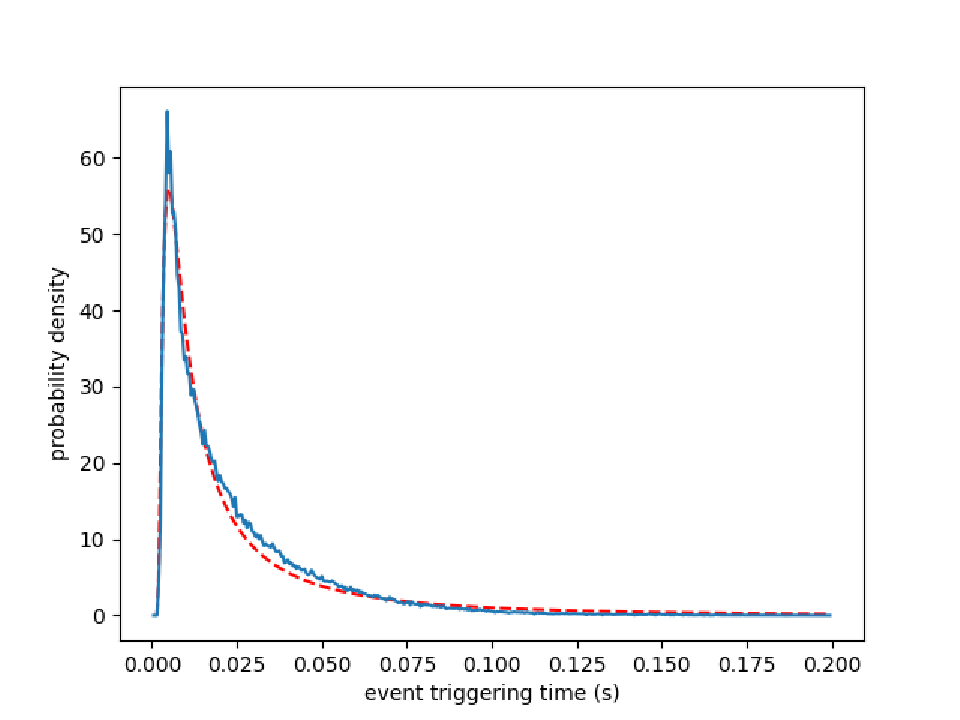}
		\caption{$\mu=500,\ L=50$}
		\label{dl500al50}
	\end{subfigure}
	\caption{distributions of event triggering time. Blue lines indicate the histograms of event triggering time and red dashed lines are the fitting curves. For different $L$, the lengths of discontinuity remain stable, demonstrating that the time delay $\Delta t_e$ is not related to current light intensity. Different from that, the changing speed $\mu$ decreases the length of discontinuity.}
	\label{fig:discontinuity subfigure}
\end{figure*}

\begin{figure}[htb]
	\centering
	\includegraphics[width=8.5cm]{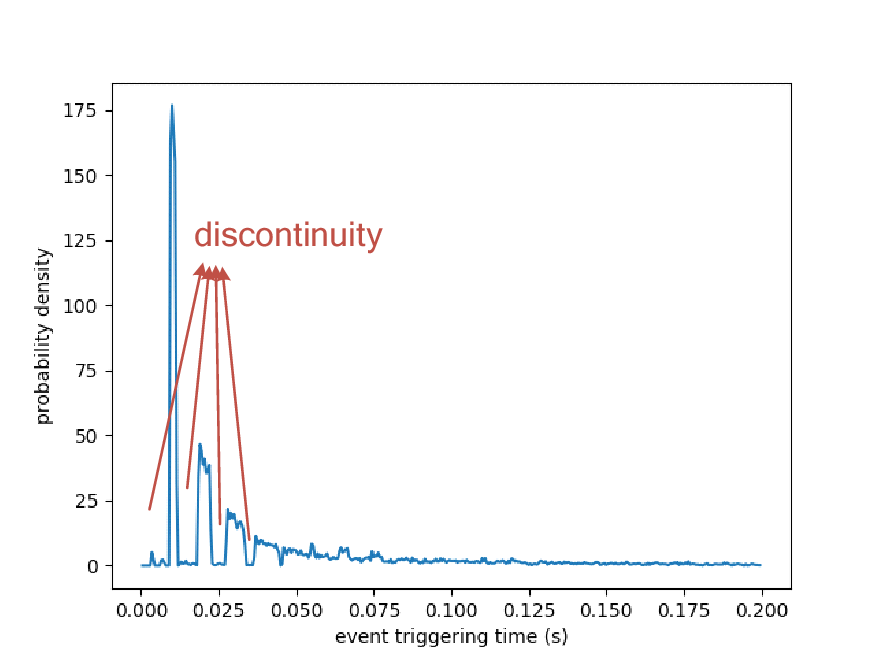}
	\caption{exemplified discontinuity of event triggering time. (Here $\mu=50$, $L=10$)}
	\label{fig:discontinuity_illustration}
\end{figure}

Fig. \ref{fig:discontinuity subfigure} shows the statistical distributions of event triggering time as the histograms with blue lines in different conditions. The fitting curves above the histograms, marked as red dashed lines, roughly follow the inverse Gaussian distribution. As is exemplified in Fig. \ref{fig:discontinuity_illustration}, there exists significant discontinuity in the histograms, which is also shown in Fig. \ref{fig:discontinuity subfigure}. It validates the existence of the time delay $\Delta t_e$ between consecutive triggering events caused by charge and discharge of parasitic capacitor of photodiode $\rm PD$.

In Fig. \ref{fig:discontinuity subfigure}, the lengths of discontinuity remain the same in each row but vary in each column, which demonstrates that the time delay $\Delta t_e$ only depends on $\mu$ rather than current light intensity $L$. Therefore, the time delay $\Delta t_e$ follows a non-first-order behavior.

\begin{table*}[htbp]
	\centering
	\caption{time delay $\Delta t_e$}
	\begin{tabular}{|c|c|c|c|c|c|c|}
		\hline
		\multicolumn{2}{|c|}{} &\multicolumn{5}{|c|}{current light intensity $L$} \\
		\cline{3-7}
		\multicolumn{2}{|c|}{}  & 10 & 20 & 30 & 40 & 50\\\hline
		\multirow{8}{*}{\rotatebox{90}{light changing speed $\mu$}}  & 50 & \(9.0*e^{-3}s\) &  \(9.0*e^{-3}s \)&  \(9.0*e^{-3}s\) &  \(9.0*e^{-3}s\) &  \(9.0*e^{-3}s\) \\\cline{2-7}    
		&60 & \(7.5*e^{-3}s\) &  \(7.5*e^{-3}s\) &  \(7.5*e^{-3}s\) &  \(7.5*e^{-3}s\) &  \(7.5*e^{-3}s\) \\\cline{2-7}
		& 70 & \(6.5*e^{-3}s\) &  \(6.5*e^{-3}s\) &  \(6.5*e^{-3}s\) &  \(6.5*e^{-3}s\) &  \(6.5*e^{-3}s\) \\\cline{2-7}
		& 80 & \(5.5*e^{-3}s\)&   \(5.5*e^{-3}s\)&   \(5.5*e^{-3}s\)    & \(5.5*e^{-3}s\) &    \(5.5*e^{-3}s\) \\ \cline{2-7}
		& 90 & \(4.5*e^{-3}s\)&   \(4.5*e^{-3}s\)&   \(4.5*e^{-3}s\)    & \(4.5*e^{-3}s\) &    \(4.5*e^{-3}s\) \\ \cline{2-7}
		& 100 & \(4.5*e^{-3}s\)&   \(4.5*e^{-3}s\)&   \(4.5*e^{-3}s\)    & \(4.5*e^{-3}s\) &    \(4.5*e^{-3}s\) \\ \cline{2-7}
		& 150 & \(3.0*e^{-3}s\)&   \(3.0*e^{-3}s\)&   \(3.0*e^{-3}s\)    & \(3.0*e^{-3}s\) &    \(3.0*e^{-3}s\) \\ \cline{2-7}
		& 200& \(2.0*e^{-3}s\) & \(2.0*e^{-3}s\)& \(2.0*e^{-3}s\)&   \(2.0*e^{-3}s\)&   \(2.0*e^{-3}s\)\\
		\hline  
	\end{tabular} \\
	\footnotemark{The time delay $\Delta t_e$ is measured on the first discontinuous region, as can be seen in Fig. \ref{fig:discontinuity_illustration}. Because it is more accurate compared with others that are influenced by the unstable changing speed of light}.\\
	\label{tbl:quantitative evaluation}
\end{table*}

We also evaluate the discontinuity length quantitatively in Table \ref{tbl:quantitative evaluation}. The time delay $\Delta t_{e}$ declines with the increasing of changing speed $\mu$, which can also be seen in Fig. \ref{fig:discontinuity subfigure}. Furthermore, the product of $\Delta t_e$ and $\mu$ is almost unchanged with the light intensity $L$. It proves our analysis of the inversely proportional relation between them.

The time delay $\Delta t_{e}$ exists with a non-first-order behavior, unlike the usual manner of first-order system of DVS. In the dim light conditions, the changing speed of light is slow because the absolute difference of light is small. As a result, the time delay $\Delta t_{e}$ is more significant.

\section{CONCLUSIONS}
\label{sec:typestyle}

In this paper, we study on the properties of DVS circuit, and find a new behavior of DVS: the time delay $\Delta t_e$ and discontinuity of event triggering time. It leads to a non-first-order behavior, different from the usual manners of DVS. The time delay is inversely proportional to the changing speed of light. In dim light conditions, the difference of light intensity is small, slowing the light variation. As a result, the time delay and discontinuity become prominent in dim light conditions. The experimental results are also provided for validation.

\bibliographystyle{IEEEbib}
\bibliography{strings,refs}

\begin{thebibliography}{10}

\bibitem{delbruck1994adaptive}
T.~Delbruck and C.A. Mead,
\newblock ``Adaptive photoreceptor with wide dynamic range,''
\newblock in {\em Proceedings of IEEE International Symposium on Circuits and
  Systems-ISCAS'94}. IEEE, 1994, vol.~4, pp. 339--342.

\bibitem{delbruck2012icip}
J.H. Lee, P.K.J. Park, C.-W. Shin, H.~Ryu, B.C. Kang, and T.~Delbruck,
\newblock ``Touchless hand gesture ui with instantaneous responses,''
\newblock in {\em 2012 IEEE International Conference on Image Processing}.
  IEEE, 2012, pp. 1957--1960.

\bibitem{delbruck2022icip}
T.~Delbruck, C.~Li, R.~Graca, and B.~Mcreynolds,
\newblock ``Utility and feasibility of a center surround event camera,''
\newblock in {\em 2022 IEEE International Conference on Image Processing}.
  IEEE, 2022, pp. 381--385.

\bibitem{delbruck2008IEEEJSSC}
Patrick Lichtsteiner, Christoph Posch, and Tobi Delbruck,
\newblock ``A 128$\times$ 128 120 db 15 $\mu$s latency asynchronous temporal
  contrast vision sensor,''
\newblock {\em IEEE Journal of Solid-State Circuits}, vol. 43, no. 2, pp.
  566--576, 2008.

\bibitem{delbruck2022IEEETPAMI}
G.~Gallego, T.~Delbrück, G.~Orchard, C.~Bartolozzi, B.~Taba, A.~Censi,
  S.~Leutenegger, A.J. Davison, J.~Conradt, K.~Daniilidis, and D.~Scaramuzza,
\newblock ``Event-based vision: A survey,''
\newblock {\em IEEE Transactions on Pattern Analysis and Machine Intelligence},
  vol. 44, no. 1, pp. 154--180, 2022.

\bibitem{mahowald1991silicon}
M.A Mahowald,
\newblock ``Silicon retina with adaptive photoreceptors,''
\newblock in {\em Visual information processing: from neurons to chips}. SPIE,
  1991, vol. 1473, pp. 52--58.

\bibitem{indiveri2011neuromorphic}
G.~Indiveri, B.~Linares-Barranco, T.J. Hamilton, Andr{\'e} Van~S.,
  R.~Etienne-Cummings, T.~Delbruck, S.-C. Liu, P.~Dudek, P.~H{\"a}fliger,
  S.~Renaud, et~al.,
\newblock ``Neuromorphic silicon neuron circuits,''
\newblock {\em Frontiers in neuroscience}, vol. 5, pp. 9202, 2011.

\bibitem{muglikar2021calibrate}
M.~Muglikar, M.~Gehrig, D.~Gehrig, and D.~Scaramuzza,
\newblock ``How to calibrate your event camera,''
\newblock in {\em Proceedings of the IEEE/CVF Conference on Computer Vision and
  Pattern Recognition}, 2021, pp. 1403--1409.

\bibitem{brewer2021comparative}
T.~Brewer and M.~Hawks,
\newblock ``A comparative evaluation of the fast optical pulse response of
  event-based cameras,''
\newblock in {\em Image Sensing Technologies: Materials, Devices, Systems, and
  Applications VIII}. SPIE, 2021, vol. 11723, pp. 86--103.

\bibitem{mueggler2017event}
E.~Mueggler, H.~Rebecq, G.~Gallego, T.~Delbruck, and D.~Scaramuzza,
\newblock ``The event-camera dataset and simulator: Event-based data for pose
  estimation, visual odometry, and slam,''
\newblock {\em The International Journal of Robotics Research}, vol. 36, no. 2,
  pp. 142--149, 2017.

\bibitem{gehrig2020video}
D.~Gehrig, M.~Gehrig, J.~Hidalgo-Carri{\'o}, and D.~Scaramuzza,
\newblock ``Video to events: Recycling video datasets for event cameras,''
\newblock in {\em Proceedings of the IEEE/CVF Conference on Computer Vision and
  Pattern Recognition}, 2020, pp. 3586--3595.

\bibitem{hu2021v2e}
Y.~Hu, S.-C. Liu, and T.~Delbruck,
\newblock ``v2e: From video frames to realistic dvs events,''
\newblock in {\em Proceedings of the IEEE/CVF Conference on Computer Vision and
  Pattern Recognition}, 2021, pp. 1312--1321.

\bibitem{graca2021unraveling}
Rui Graca and Tobi Delbruck,
\newblock ``Unraveling the paradox of intensity-dependent dvs pixel noise,''
\newblock {\em arXiv preprint arXiv:2109.08640}, 2021.

\bibitem{lin2022dvs}
S.~Lin, Y.~Ma, Z.~Guo, and B.~Wen,
\newblock ``Dvs-voltmeter: Stochastic process-based event simulator for dynamic
  vision sensors,''
\newblock in {\em European Conference on Computer Vision}. Springer, 2022, pp.
  578--593.

\bibitem{dentan1990numerical}
M.~Dentan and B.~de~Cremoux,
\newblock ``Numerical simulation of the nonlinear response of a pin photodiode
  under high illumination,''
\newblock {\em Journal of Lightwave Technology}, vol. 8, no. 8, pp. 1137--1144,
  1990.

\bibitem{razavi2021fundamentals}
B.~Razavi,
\newblock {\em Fundamentals of microelectronics},
\newblock John Wiley \& Sons, 2021.

\bibitem{gracca2023shining}
R.~Gra{\c{c}}a, B.~McReynolds, and T.~Delbruck,
\newblock ``Shining light on the dvs pixel: A tutorial and discussion about
  biasing and optimization,''
\newblock in {\em Proceedings of the IEEE/CVF Conference on Computer Vision and
  Pattern Recognition}, 2023, pp. 4044--4052.

\bibitem{jiang2018super}
H.~Jiang, D.~Sun, V.~Jampani, M.-H. Yang, E.~Learned-Miller, and J.~Kautz,
\newblock ``Super slomo: High quality estimation of multiple intermediate
  frames for video interpolation,''
\newblock in {\em Proceedings of the IEEE conference on computer vision and
  pattern recognition}, 2018, pp. 9000--9008.

\bibitem{rebecq2018esim}
H.~Rebecq, D.~Gehrig, and D.~Scaramuzza,
\newblock ``Esim: an open event camera simulator,''
\newblock in {\em Conference on robot learning}. PMLR, 2018, pp. 969--982.

\bibitem{series2017colour}
B~Series,
\newblock ``Colour gamut conversion from recommendation itu-r bt. 2020 to
  recommendation itu-r bt. 709,''
\newblock {\em International Telecommunication Union}, 2017.

\end{thebibliography}

\end{document}